\documentclass[sn-basic]{sn-jnl}

\usepackage{graphicx}%
\usepackage{multirow}%
\usepackage{amsmath,amssymb,amsfonts}%
\usepackage{amsthm}%
\usepackage{mathrsfs}%
\usepackage[title]{appendix}%
\usepackage{xcolor}%
\usepackage{textcomp}%
\usepackage{manyfoot}%
\usepackage{booktabs}%
\usepackage[linesnumbered,ruled,vlined]{algorithm2e}
\usepackage{algpseudocode}%
\usepackage{listings}%
\usepackage{adjustbox}%

\theoremstyle{thmstyleone}%
\theoremstyle{thmstyletwo}%

\theoremstyle{thmstylethree}%

\begin{document}

\title[Article Title]{SocialTrack: Multi-Object Tracking in Complex Urban Traffic Scenes Inspired by Social Behavior}


\author[1]{\fnm{Wenguang} \sur{Tao}}\email{wenguangtao2022@mail.nwpu.edu.cn}
\equalcont{These authors contributed equally to this work.}

\author*[1]{\fnm{Xiaotian} \sur{Wang}}\email{xtwang0111@nwpu.edu.cn}
\equalcont{These authors contributed equally to this work.}

\author[1]{\fnm{Tian} \sur{Yan}}\email{tianyan@nwpu.edu.cn}
\equalcont{These authors contributed equally to this work.}

\author[1]{\fnm{Jie} \sur{Yan}}\email{jyan@nwpu.edu.cn}
\equalcont{These authors contributed equally to this work.}

\author[1]{\fnm{Guodong} \sur{Li}}\email{liguodong23@mail.nwpu.edu.cn}
\equalcont{These authors contributed equally to this work.}

\author[2]{\fnm{Kun} \sur{Bai}}\email{baikundb@126.com}
\equalcont{These authors contributed equally to this work.}

\affil*[1]{\orgdiv{Unmanned Systems Research Institute}, \orgname{Northwestern Polytechnical University}, \orgaddress{\city{Xi'an}, \postcode{710072}, \state{Shaanxi}, \country{China}}}

\affil[2]{\orgname{Xi'an Modern Control Technology Research Institute}, \orgaddress{\city{Xi'an}, \postcode{710065}, \state{Shaanxi}, \country{China}}}


\abstract{As a key research direction in the field of multi-object tracking (MOT), UAV-based multi-object tracking has significant application value in the analysis and understanding of urban intelligent transportation systems. However, in complex UAV perspectives, challenges such as small target scale variations, occlusions, nonlinear crossing motions, and motion blur severely hinder the stability of multi-object tracking. To address these challenges, this paper proposes a novel multi-object tracking framework, SocialTrack, aimed at enhancing the tracking accuracy and robustness of small targets in complex urban traffic environments. The specialized small-target detector enhances the detection performance by employing a multi-scale feature enhancement mechanism. The Velocity Adaptive Cubature Kalman Filter (VACKF) improves the accuracy of trajectory prediction by incorporating a velocity dynamic modeling mechanism. The Group Motion Compensation Strategy (GMCS) models social group motion priors to provide stable state update references for low-quality tracks, significantly improving the target association accuracy in complex dynamic environments. Furthermore, the Spatio-Temporal Memory Prediction (STMP) leverages historical trajectory information to predict the future state of low-quality tracks, effectively mitigating identity switching issues. Extensive experiments on the UAVDT and MOT17 datasets demonstrate that SocialTrack outperforms existing state-of-the-art (SOTA) methods across several key metrics. Significant improvements in MOTA and IDF1, among other core performance indicators, highlight its superior robustness and adaptability. Additionally, SocialTrack is highly modular and compatible, allowing for seamless integration with existing trackers to further enhance performance.}

\keywords{Multi-object tracking, spatio-temporal memory fusion, multi-scale context aggregation, group motion compensation, UAV}



\maketitle

\section{Introduction}\label{sec1}

As one of the core tasks in the field of machine vision, multi-object tracking (MOT) \citep{1vandenhende2021multi} holds significant importance in the context of urban intelligent transportation systems. The main objective is to maintain continuous motion trajectories for each detected object in a video sequence \citep{2nasir2025enhanced}. In urban traffic management, MOT technology not only requires efficient object detection but also necessitates powerful object association capabilities to address challenges \citep{fu2023siamese} such as occlusions, object crossings, and rapid motion in complex environments. With the rapid development of computer vision, hardware capabilities, and sensor technologies, MOT has shown immense potential in applications such as autonomous driving \citep{3buyval2018realtime, 4wang2023camo}, urban traffic monitoring \citep{5jiao2024digital}, intelligent transportation systems \citep{7nguyen2023multi}, and urban planning.

In urban traffic scenarios, the application of MOT spans a wide range, including the detection and tracking of vehicles, pedestrians, roads, and other elements. For example, in intelligent transportation systems, MOT technology can help with dynamic traffic flow monitoring, road congestion analysis, and traffic trend prediction \citep{6fei2023multi}. With the rapid development of unmanned aerial vehicle (UAV) technology \citep{8mohaimenianpour2018hands}, UAVs, with their high maneuverability, convenient deployment, and flexible viewpoints, have become important tools for urban traffic monitoring \citep{9zhu2021detection, 10liu2023robust}. UAVs equipped with cameras can be used for monitoring urban traffic conditions \citep{11mohsan2022towards}, detecting traffic accidents, and assessing road safety \citep{12li2023high}. However, the complex urban environment, especially highly dynamic traffic scenarios, presents new challenges for object tracking \citep{13guan2025multi}, such as target scale variation, occlusions, nonlinear crossing motions, and motion blur.

To address these challenges, most existing MOT methods adopt a tracking-by-detection (TBD) paradigm \citep{14ariza2024object}. The core idea is to independently detect objects in each frame using an object detector, and then use association algorithms to match the detected objects across frames, thus forming continuous motion trajectories. This approach effectively improves the accuracy and real-time performance of multi-object tracking. However, the complex dynamic characteristics of urban traffic environments, particularly the drastic viewpoint changes and unstable motion postures encountered by UAVs during flight, pose challenges to traditional Kalman filter \citep{15zhong2024kalman}. Traditional filtering methods struggle to accurately predict the position of the objects' trajectories, leading to significant trajectory misalignment and discontinuities, which affect the stability and accuracy of object tracking in urban traffic scenarios.

To address the issue of trajectory instability, recent research has primarily focused on appearance learning and association design \citep{16hassan2024multi}. In terms of appearance learning, MeMOT \citep{17cai2022memot} retains a large spatiotemporal memory to store the identity embeddings of the tracked targets, adaptively referencing and aggregating useful information from memory as needed. The work in \citep{18ren2023focus} explores diversified fine-grained representations of target appearance using high feature resolution and precise semantic information. Although these methods enhance the discriminability of appearance features, they are ineffective in handling small targets with missing appearance features and require additional computational resources, which hinders real-time tracking. In terms of association design, the work in \citep{19seidenschwarz2023simple} combines appearance features with a simple motion model to achieve strong tracking results. SparseTrack \citep{20liu2025sparsetrack} proposes a pseudo-depth estimation method for deep cascading matching to improve occlusion handling in target association. BoT-SORT \citep{21aharon2022bot} employs dedicated camera motion compensation (CMC) technology \citep{jiao2023survey} to achieve frame alignment and handle dynamic scenes. Similarly, OC-SORT \citep{22cao2023observation} constructs virtual trajectory segments during occlusions based on target observations to fill in missing trajectory information. Although these association strategies are effective for multi-object tracking, they are often designed for specific datasets or require fine-tuning of hyperparameters for particular scenarios.

Unlike existing methods, this paper proposes a framework for multi-object tracking of small targets in complex urban traffic environments-SocialTrack. This framework integrates several strategies, including detector improvements, trajectory filtering, group prior modeling, and spatiotemporal prediction, to systematically enhance the tracking performance of small targets under challenging conditions such as occlusions, dense scenes, and motion blur. First, we introduce a multi-scale feature enhancement mechanism and a context encoding-decoding perception module into the detector architecture, called SOFEPNet, which effectively improves the response strength of small targets in the feature map. Next, to improve the accuracy of trajectory state estimation, we propose the Velocity Adaptive Cubature Kalman Filter (VACKF). This method incorporates a velocity dynamic modeling mechanism into the traditional filter framework and uses a nonlinear volumetric representation of the state space to improve the modeling of fast-moving small targets and nonlinear trajectory variations. To further enhance performance in dense object environments, we introduce a Group Motion Compensation Strategy (GMCS). By constructing a motion consistency modeling network in the local neighborhood, it provides stable motion references for low-quality trajectories, enabling state compensation and correction. Finally, we propose a Spatio-Temporal Memory Prediction (STMP) module, which is specifically designed to infer the future state of low-quality trajectories by leveraging the target's own historical trajectory information in the absence of external observations or when the group compensation fails. Experimental results on the UAVDT and MOT17 public datasets demonstrate that the proposed SocialTrack outperforms existing state-of-the-art methods in terms of performance.

In summary, the contributions of this paper are as follows:

\begin{enumerate}
	\item{A new framework for multi-object tracking of small targets in complex environments, SocialTrack, is proposed. This framework integrates several innovative strategies to effectively address the tracking challenges of small targets under conditions such as occlusions, dense scenes, and motion blur. SocialTrack not only achieves state-of-the-art (SOTA) performance but also offers high compatibility, allowing for seamless integration with other trackers.}
	\item{SOFEPNet is designed to enhance the feature response strength of small targets in the object detector, significantly improving detection performance, especially in complex backgrounds and low-quality detection scenarios.}
	\item{AVCKF is proposed to improve the modeling of fast-moving small targets and nonlinear trajectory variations by introducing a velocity dynamic modeling mechanism and nonlinear volumetric representation, thus enhancing the accuracy and robustness of trajectory state estimation.}
	\item{GMCS is designed to model motion consistency in the local neighborhood, providing stable motion references for low-quality trajectories. This strategy effectively reduces identity switching and target loss, particularly improving target association accuracy in complex dynamic environments.}
	\item{STMP is introduced to infer the future state of low-quality trajectories using historical trajectory information in the absence of external observations or when group compensation fails, further optimizing the continuity and stability of targets and enhancing the system's adaptability to missing observations.}
\end{enumerate}

The rest of the paper is organized as follows: Section \ref{section_related_work} discusses related work on tracking frameworks; Section \ref{section_methodology} presents the overall architecture of the SocialTrack and the proposed improvement strategies; Section \ref{section_experiment} provides comprehensive experimental validation of the proposed improvements; and Section \ref{conclusion} concludes the paper.

\section{Related Work}\label{section_related_work}

\subsection{MOT}

Multi-object tracking algorithms are generally classified into two categories: two-stage and single-stage methods. Two-stage methods primarily include tracking-by-detection (TBD), while single-stage methods encompass joint detection and tracking (JDT) as well as transformer-based tracking approaches.

\noindent {\bf{Two-stage.}} Two-stage object trackers first perform the detection task and then independently execute the tracking task based on the detection results. SORT \citep{23bewley2016simple} is one of the most classic algorithms under the tracking-by-detection (TBD) paradigm. It predicts the future position of the trajectory using Kalman filter and matches the detection results with the trajectories via the Hungarian algorithm \citep{24kuhn1955hungarian}. DeepSORT \citep{25wojke2017simple} extends SORT by introducing re-identification features and enhancing the tracker's performance in scenes with low motion distinction through the fusion of appearance features. ByteTrack \citep{26zhang2022bytetrack} improves tracking performance in occlusion scenarios by performing secondary matching of low-confidence detection results with high-quality tracking trajectories. To address the distortion caused by rapid viewpoint changes affecting the filter's prediction, BoT-SORT \citep{21aharon2022bot} corrects the bounding box position by generating a camera motion compensation matrix for each frame. OC-SORT \citep{22cao2023observation} focuses on target-centric adjustment strategies and proposes a target center matching method that significantly improves the tracking of nonlinear moving targets. The C-BIoU \citep{27yang2023hard} tracker adds a buffer to expand the matching space between detection results and trajectories. LTTrack \citep{28lin2024lttrack} encodes relative and absolute positions as interaction and motion features for association, significantly reducing identity switches in long-term tracking. However, previous studies often face challenges in handling rapid motion in UAV videos. Co-MOT \citep{29liu2025co} introduces Graph Neural Networks (GNN) \citep{30han2022vision} into the tracking algorithm to simulate interactions between targets, achieving significant improvements. MCCA-MOT \citep{31li2024mcca} designs a cascading strategy for targets with blurred features, applying different similarity metrics at different association stages to reduce identity switches and trajectory fragmentation.

\noindent {\bf{Single-stage.}} Single-stage object trackers perform both detection and motion/appearance feature extraction simultaneously, ultimately outputting the target's position and unique identity. CenterTrack \citep{32zhou2020tracking} focuses on the center of the target, directly predicting the target's position and motion offset using a network. TraDeS \citep{33wu2021track} uses tracking clues to end-to-end assist detection, constructing a cost matrix to infer motion offsets for target tracking and propagating previous target features based on offsets to improve detection performance. In terms of appearance feature information, Swin-JDE \citep{34tsai2023swin} employs neural network learning for upsampling, enhancing the spatial information of the feature map so that a single network can simultaneously perform detection and re-identification tasks. FairMOT \citep{35zhang2021fairmot} introduces a re-identification branch to predict target features and jointly optimizes the detection and re-identification tasks through a carefully designed network structure. Chained-Tracker \citep{36peng2020chained} integrates object detection, feature extraction, and data association into a single end-to-end solution for the first time. HomView-MOT \citep{37ji2024view} incorporates homographic matching and viewpoint-centric concepts, achieving cross-view target association through multi-view homogeneity learning. U2MOT \citep{38liu2023uncertainty} improves feature consistency by leveraging uncertainty, proposing an unsupervised multi-object tracking framework based on uncertainty metrics and trajectory segment-guided enhancement strategies. UCMCTrack \citep{39yi2024ucmctrack} effectively handles challenges caused by camera motion by applying unified camera motion compensation parameters throughout the video sequence and introducing position mapping. TrackFormer \citep{40meinhardt2022trackformer} uses attention mechanisms to autoregressively track target trajectories in video sequences, without requiring additional graph optimization or appearance modeling. MOTR \citep{41zeng2022motr} extends DETR \citep{42carion2020end} by introducing "tracking queries" and combines small trajectory-aware label assignment with a temporal aggregation network to effectively model temporal relationships and improve tracking accuracy. TrackingMamba \citep{43wang2024trackingmamba} adopts a single-stream architecture with a Vision Mambacit \citep{44yu2025mambaout} backbone, combining global feature extraction and long-range dependency modeling, significantly reducing computational overhead.
\begin{algorithm}[!b]
	\caption{Pseudo-code of SocialTrack.}
	\label{alg:tracking algorithm}
	\SetAlgoLined
	
	\SetKwInOut{Input}{Input}
	\SetKwInOut{Output}{Output}
	
	\Input{Video sequence $V$; object detector $SOFEPNet$; detection score threshold $\tau$}
	\Output{Tracks $T$ of the video $V$}
	
	\BlankLine
	Initialization: $T$ $\leftarrow \emptyset$ \\
	\For{frame $f_k$ in $V$}{
		
		\begingroup
		\color{brown!70!black}
		/\# Small Object Feature Efficient Perception Detection in \textbf{Section 3.B} \#/\\
		$D_k$ $\leftarrow$ $SOFEPNet({f_k})$ \tcp{detections per frame}
		$D_{high}$, $D_{low}$ $\leftarrow$ $D_k$ \tcp{divided into two parts by $\tau$}
		\endgroup
		\BlankLine
		
		\tcc{1st association for $T_{high}$}
		\BlankLine
		
		\begingroup
		\color{blue}
		/\# Velocity Adaptive Cubature Kalman Filter in \textbf{Section 3.C} \#/\\
		$T \leftarrow VACKF\_predict(T)$ \\
		Associate $T$ and $D_{high}$ use Hungarian Algorithm: \\
		${T_{low}},{T_{high}} \leftarrow HM(T,{D_{high}})$ \\
		\For{$t$ in ${T_{high}}$}
		{
			$t \leftarrow VACKF\_update(t)$ \\
		}
		\endgroup
		\BlankLine
		
		\tcc{2st association for $T_{low}$}
		\BlankLine
		
		\begingroup
		\color{teal}
		/\# Group Motion Compensation Strategy in \textbf{Section 3.D} \#/\\
		\tcp{Using GMCS to update status}
		\For{$t$ in ${T_{high}}$}{
			\If{$t$ similarity with $T_{low}$ $>$ $threshold$}{
				${T_{neighbor}} \leftarrow {T_{neighbor}} \cup \{ t\}$}
		}
		update the $T_{low}$ status use ${T_{neighbor}}$: \\
		$T_{low}^{center} = {{\sum\limits_{i = 1}^N {{T_{p,i}}}  \cdot S(T,{\mkern 1mu} {T_i}^H)} \mathord{\left/
				{\vphantom {{\sum\limits_{i = 1}^N {{T_{p,i}}}  \cdot S(T,{\mkern 1mu} {T_i}^H)} {\sum\limits_{i = 1}^N {S(T,{\mkern 1mu} {T_i}^H)} }}} \right.
				\kern-\nulldelimiterspace} {\sum\limits_{i = 1}^N {S(T,{\mkern 1mu} {T_i}^H)} }}$
		${T_{low}} \leftarrow VACKF\_update(T_{low}^{center})$
		\endgroup
		\BlankLine
		
		\begingroup
		\color{purple}
		\# Spatio-Temporal Memory Prediction in \textbf{Section 3.E} \#\\
		\tcp{GMCS is disabled at this moment. Using STMP to update status}
		\If{${T_{neighbor}} = 0$ }{
			$ T_{low}^{center} = Cascade\_LSTM({T_{low}})$
			${T_{low}} \leftarrow VACKF\_update(T_{low}^{center})$
		}
		\endgroup
	}
	\Return{$T$}
\end{algorithm}

\subsection{Data association}

Data association is a critical step in Multi-Object Tracking (MOT). It works by evaluating the similarity between existing trajectories and the detection results of the current frame, thereby matching targets across different frames and assigning a unique ID to each target \citep{45emami2020machine}. \cite{bi2025linear} propose a linear projection method based on graph convolutional networks for effective correlation and classification of data. MOTDT \citep{46chen2018real} collects candidate targets from both detection and tracking outputs and utilizes a scoring function based on fully convolutional neural networks to select the optimal target for matching in real-time. QDTrack \citep{47pang2021quasi} proposes a quasi-dense similarity learning method by densely sampling hundreds of region proposals on a pair of images for contrastive learning, and performs simple nearest-neighbor search using the obtained feature space. SparseTrack \citep{20liu2025sparsetrack} improves data association performance in crowded and occluded scenarios by using pseudo-depth estimation and the Deep Cascade Matching (DCM) strategy, which transforms dense target sets into sparse target subsets. Unlike the aforementioned methods, we leverage group motion pattern learning to infer the reliable motion states of all targets and apply this knowledge to existing trajectories.

\section{Methodology}\label{section_methodology}

\subsection{Overview of the Framework}

In this section, we introduce SocialTrack, a strategy specifically designed to address small target tracking in complex UAV perspectives. SocialTrack follows the tracking-by-detection paradigm and establishes trajectories in continuous video sequences through data association, with the detailed implementation steps outlined in Algorithm \ref{alg:tracking algorithm}. SocialTrack takes the sequence $V$ as input, first utilizing a small object feature efficient perception network (SOFEPNet) to predict bounding boxes and confidence scores in complex environments. Then, high-confidence detection results are matched with existing trajectories, and Velocity Adaptive Cubature Kalman Filter (VACKF) is applied to update the states of the matched trajectories.

\begin{figure}[!h]
	\centering
	\includegraphics[width=\textwidth]{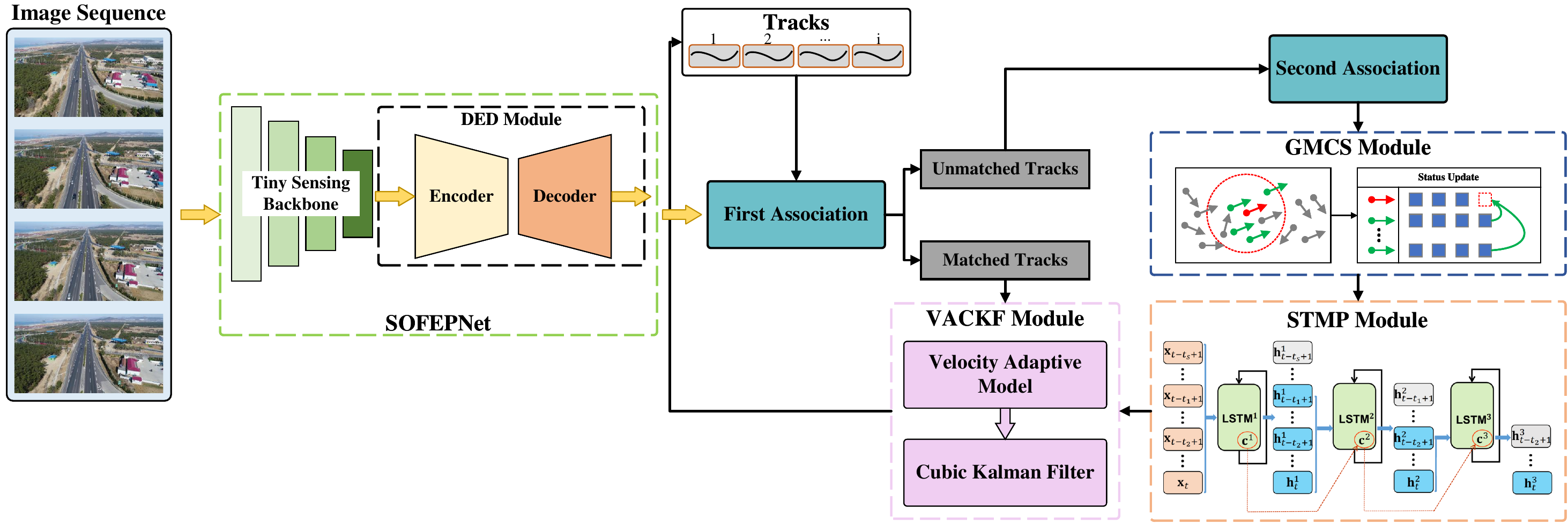}
	\caption{The overall architecture of SocialTrack. For the input image sequence, SOFEPNet first achieves accurate detection of small targets. Subsequently, VACKF performs state updates for the matched trajectories. For unmatched trajectories, GMCS predicts the trajectory states using group motion priors. If GMCS fails, STMP is activated to predict the object positions at the next moment based on historical trajectories.
	}
	\label{fig_1}
\end{figure}

Unlike existing methods such as \citep{23bewley2016simple, 25wojke2017simple, 26zhang2022bytetrack} that focus solely on high-quality trajectories, we allocate more resources to handling low-quality tracking trajectories, as these are the main causes of identity switches and trajectory drift. For unmatched low-quality trajectories, we design a Group Motion Compensation Strategy (GMCS) as prior knowledge to update the trajectory states, preventing error accumulation. When the motion compensation strategy fails, a Spatio-Temporal Memory Prediction (STMP) module is employed to predict the object's position at the next time step using historical trajectory information. The final output of the algorithm is a set of trajectories $T$, where each trajectory includes cross-frame bounding box coordinates and the unique ID of the object. The overall architecture of SocialTrack is demonstrated in Fig. \ref{fig_1}.

\subsection{Small Object Feature Efficient Perception Network}

UAVs typically operate at high altitudes, with onboard cameras providing wide fields of view. However, the targets captured in the images are often small, with low resolution, and lack shape, structure, and texture information \citep{hua2025survey}. Moreover, UAVs encounter various shooting environments, such as changes in viewpoint (e.g., top-down, side-view), lighting conditions (e.g., sunlight, night), and weather conditions (e.g., fog, rain). Conventional object detection methods utilize backbone networks like ResNet \citep{48he2016deep} to extract features at different scales, then perform simple multi-scale feature fusion using feature pyramid networks, ultimately producing detection results at different scales to tackle multi-scale object detection tasks. However, these networks face several issues when handling small targets: 1) excessive downsampling of high-level features results in a significant loss of crucial information for small targets; 2) low-level features lack semantic information, leading to inadequate representation of small target features; 3) inefficient feature fusion mechanisms prevent the effective enhancement of small target features.

The key to accurately detecting small targets in complex UAV perspectives lies in efficient feature extraction and fusion. For small targets, detailed information is more critical than contextual information. Shallow features from the early layers of the backbone network, due to their high resolution, contain rich detailed information that is vital for recognizing and localizing small targets. To make the network more focused on small targets, we modify the overall structure of the backbone network, allocating more computational resources to the earlier stages of the network to extract rich content information from high-resolution features of small targets. Additionally, we replace traditional 2D convolutions with dynamic convolutions to suppress background interference and enhance the structural edges of small targets. Moreover, as the network depth increases, it contains richer abstract feature representations, which are often lacking in shallow features.

\begin{figure}[!h]
	\centering
	\includegraphics[width=\columnwidth]{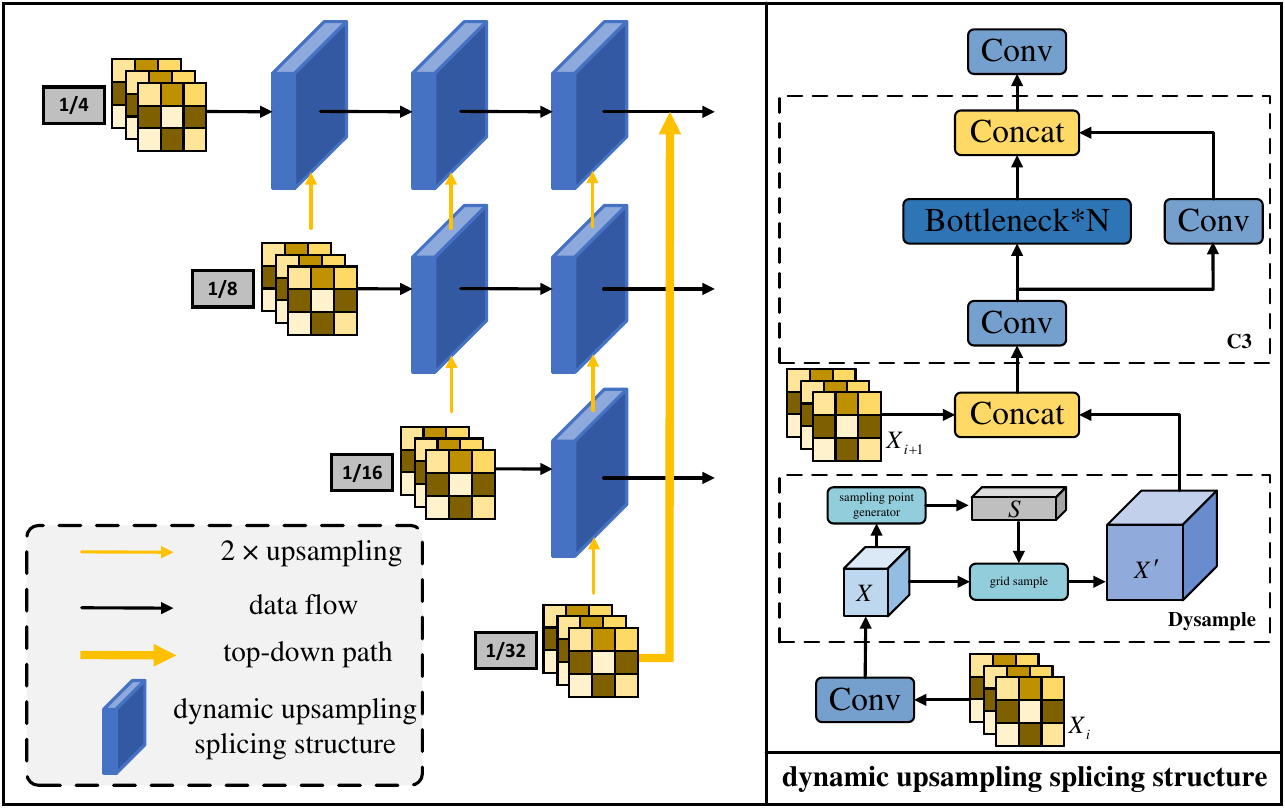}
	\caption{Structure of DED module.
	}
	\label{fig_2}
\end{figure}

An effective approach is to fuse semantic and structural texture information to enhance feature granularity and discriminability. Inspired by the encoder-decoder network U-Net \citep{49ronneberger2015u} and the lightweight dynamic upsampling module DySample \citep{50liu2023learning}, we propose a novel dynamic encoder-decoder structure (DED). It replaces the feature pyramid network (FPN) for lossless fusion of features at different scales, as shown in Fig. \ref{fig_2}. Unlike U-Net, we establish richer cross-layer connections between low-level and high-level features and replace all simple interpolation-based upsampling with DySample to address information loss and feature misalignment during the fusion process. DED takes features from layers 2-5 after downsampling by the backbone network and outputs high-level mixed enhanced features from layers 2-4, which are then fed into the detection head.

Feature upsampling is a critical step in restoring feature resolution in dense prediction models. Specifically, DED mainly consists of a dynamic upsampling splicing structure (DUSS), as shown on the right side of Fig. \ref{fig_2}. DUSS is composed of a $3 \times 3$ convolution, DySample, and the C3 module. Given a feature, after passing through a regular convolution, it is input into DySample for dynamic upsampling of feature maps, resulting in an upsampled feature of size:
\begin{equation}
	\label{eq_1}
	\begin{array}{l}
		X = {f_{3 \times 3}}({X_i})\\
		O = {f_{linear}}(X)\\
		S = G + O\\
		X' = {f_{grid\_sample}}(X,S)
	\end{array}
\end{equation}

\noindent{where ${f_{3 \times 3}}$ is the regular convolution operation; ${f_{linear}}$ is the linear layer used to generate the bias; $G$ is the original sampling grid; ${f_{grid\_sample}}$ is the built-in grid sampling function in PyTorch. Then, $X'$ is concatenated with ${X_{i + 1}} \in {\mathbb{R}^{C \times 2W \times 2H}}$ from the previous layer along the channel dimension and input into the C3 module for mixed feature extraction:}
\begin{equation}
	\label{eq_2}
	\begin{array}{l}
		{X_{out}} = {C_3}({f_{Concat}}(X',{X_{i + 1}}))
	\end{array}
\end{equation}

Based on point sampling design, DUSS reconstructs the semantic feature upsampling process by dynamically generating the sampling matrix needed for upsampling through content-aware offsets. The high-level complex abstract information flows throughout the network from top to bottom, guiding the low-level features, achieving effective fusion and representation of small target contextual information and detailed features.

\subsection{Velocity Adaptive Cubature Kalman Filter}

Due to the changes in posture and viewpoint during aerial flight, there is significant uncertainty in the UAV's motion state. Meanwhile, ground targets do not follow simple linear motion. The combination of these two factors results in complex nonlinear motion properties for the target trajectories in image sequences, presenting new and challenging issues for stable multi-object tracking. Traditional object tracking algorithms assume that targets move at a constant linear speed over short periods and use standard Kalman filter to predict the object's position and state at the next time step. Therefore, when the target undergoes complex motion, the covariance estimates provided by the standard Kalman filter suffer from significant errors and cannot provide an accurate state estimate for the object at the next time step. When observations are missing, the errors accumulate further, causing the tracking task to fail. To address this, we combine a velocity-adaptive motion model with a Cubature Kalman Filter (CKF) model and propose VACKF.

In BoT-SORT \citep{21aharon2022bot}, the state vector of the object is represented as $x = [{x_c},{y_c},w,h,{v_x},{v_y},\dot w,\dot h]$. To account for different motion states, we add acceleration $a$ as the ninth state, resulting in a new state vector $x = [{x_c},{y_c},w,h,{v_x},{v_y},\dot w,\dot h,a]$. The velocity-adaptive motion model dynamically adjusts the target's motion mode by introducing the acceleration parameter, thereby more accurately describing varying-speed motion scenarios. The model automatically switches between two motion modes based on the current acceleration: when the acceleration magnitude is below  , the constant velocity model is used; when the acceleration magnitude exceeds the threshold, an acceleration decay mechanism is applied. The acceleration formula is as follows:
\begin{equation}
	\label{eq_3}
	{a_{k + 1}} = \left\{ {\begin{array}{*{20}{c}}
			0&{{a_k} \le 0.1}\\
			{{a_k} \cdot {e^{ - 0.1\Delta t}}}&{{a_k} > 0.1}
	\end{array}} \right.
\end{equation}

Building on this, we design a CKF that is suitable for the new state vector. CKF is a nonlinear filtering algorithm that approximates the mean and covariance under nonlinear transformations using a set of deterministic volume points, which helps to handle nonlinear systems in state estimation problems \citep{51arasaratnam2009cubature}. The CKF consists of three parts: state prediction, gain calculation, and state update. The state prediction step is used to calculate the predicted state and covariance:
\begin{equation}
	\label{eq_4}
	\begin{array}{l}
		{\cal X}_{i,k|k - 1}^* = f({{\cal X}_{i,k - 1}})\\
		{{\hat x}_k}^\prime  = \frac{1}{{2n}}\sum\limits_{i = 1}^{2n} {{\cal X}_{i,k|k - 1}^*} \\
		{P_k}^\prime  = \frac{1}{{2n}}\sum\limits_{i = 1}^{2n} {{\cal X}_{i,k|k - 1}^*} {({\cal X}_{i,k|k - 1}^*)^T} - {{\hat x}_k}^\prime {({{\hat x}_k}^\prime )^T} + {Q_k}
	\end{array}
\end{equation}

\noindent{where ${{\cal X}_{i,k - 1}}$ is the generated cubature points, $f$ is the nonlinear state transition function, ${\hat x_k}^\prime$ is the predicted state, ${P_k}^\prime$ is the predicted covariance, and ${Q_k}$ is the process noise covariance. The gain calculation step is used to compute the Kalman gain:}
\begin{equation}
	\label{eq_5}
	\begin{array}{l}
		{P_{zz}} = \frac{1}{{2n}}\sum\limits_{i = 1}^{2n} {{{\cal Z}_{i,k}}} {\cal Z}_{i,k}^T - {{\hat z}_k}\hat z_k^T + {R_k}\\
		{P_{xz}} = \frac{1}{{2n}}\sum\limits_{i = 1}^{2n} {{{\cal X}_{i,k}}^\prime } {\cal Z}_{i,k}^T - {{\hat x}_k}^\prime \hat z_k^T\\
		{K_k} = {P_{xz}}P_{zz}^{ - 1}
	\end{array}
\end{equation}

\noindent{where ${{\cal Z}_{i,k}}$ is the predicted observation, ${\hat z_k}$ is the mean of the predicted observation, ${\mathcal{X}_{i,k}}^\prime$ is the new cubature point, ${P_{zz}}$ is the observation covariance, ${P_{xz}}$ is the state-observation cross-covariance, and ${K_k}$ is the Kalman gain. The state update step is used to calculate the updated state:}
\begin{equation}
	\label{eq_6}
	\begin{gathered}
		{{\hat x}_k} = {{\hat x}_k}^\prime  + {K_k}({z_k} - {{\hat z}_k}) \hfill \\
		{P_k} = {P_k}^\prime  - {K_k}{P_{zz}}K_k^T \hfill \\ 
	\end{gathered}
\end{equation}

VACKF better adapts to the non-uniform motion of objects in UAV perspectives, while preserving the CKF's capability to handle nonlinear systems.

\subsection{Group Motion Compensation Strategy}
\begin{figure}[!h]
	\centering
	\includegraphics[width=\columnwidth]{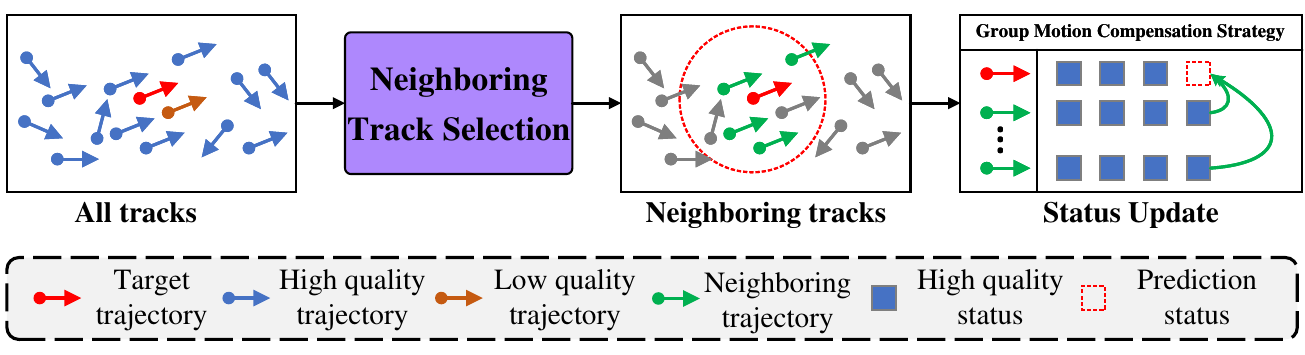}
	\caption{Structure of GMCS module.
	}
	\label{fig_3}
\end{figure}

Maintaining continuous trajectories for each target in a video sequence is one of the core tasks of multi-object tracking. In complex scenes, target occlusion is severe, and trajectories often cross and overlap. Traditional tracking methods model the motion or appearance of each object separately, relying solely on the object's own historical information to infer the state at the next time step. This approach is prone to ID switching and trajectory drift, making long-term stable tracking difficult. By analyzing trajectory segments of multiple objects in the sequence, we observed that objects of the same class and similar positions in a social group tend to exhibit very similar motion patterns. Inspired by this, we propose a group motion compensation strategy. During the tracking process, we retain the historical position and velocity information of all trajectories. When low-quality trajectories appear in the current frame (Not matched during the first association), we use the state of nearby high-quality trajectories (Matched during the first association) as prior knowledge to update the state of the low-quality trajectory in the next frame. This strategy helps avoid error accumulation from consecutive predictions, as shown in Fig. \ref{fig_3}.

Specifically, first, the similarity between low-quality trajectories and high-quality trajectories is calculated using the standardized Euclidean distance and velocity cosine distance:
\begin{equation}
	\label{eq_7}
	\begin{gathered}
		{D_{sim}} = \frac{{\left\| {{T_p} - T_p^H} \right\|}}{w} \hfill \\
		{V_{sim}} = 2 - \frac{{{T_v} \cdot T_v^H}}{{\left\| {T_v^{}} \right\| \cdot \left\| {T_v^H} \right\|}} \hfill \\ 
	\end{gathered} 
\end{equation}

\noindent{where ${T_p}$ and ${T_v}$ represent the position and velocity of the low-quality trajectory, respectively; $w$ is the width of the low-quality object; $T_p^H$ and $T_v^H$ stand for the position and velocity of the high-quality trajectory, respectively. Then, by comprehensively considering the relative distance and relative velocity, appropriate neighboring trajectories are selected for the low-quality trajectory:}
\begin{equation}
	S(T,{\mkern 1mu} {T^H}) = \left\{ {\begin{array}{*{20}{l}}
			{ - \infty }&{V \ge 2}\\
			{\frac{1}{{{D_{sim}} \cdot {V_{sim}}}}}&{V < 2}
	\end{array}} \right. 
\end{equation}

When $S(T,{\mkern 1mu} {T^H})$ is greater than the given similarity threshold, trajectory ${T^H}$ is labeled as a neighboring trajectory segment of target trajectory $T$. Finally, the prior knowledge of the group motion pattern (GMP) indicates that objects with similar states within the same group typically exhibit fixed relative velocities. The relative velocity from the previous frame can be used to infer the target's motion, providing an accurate position for the low-quality target trajectory to update the filter:
\begin{equation}
	\begin{array}{l}
		{T_{p,i}} = {{\bar T}_p} + \left( {{{\bar T}_v} - \bar T_{v,i}^H} \right) + T_{v,i}^H\\
		{T^c} = \frac{{\sum\limits_{i = 1}^N {{T_{p,i}}}  \cdot S(T,{\mkern 1mu} {T_i}^H)}}{{\sum\limits_{i = 1}^N {S(T,{\mkern 1mu} {T_i}^H)} }}
	\end{array}
\end{equation}

\noindent{where ${\bar T_p}$ and ${\bar T_v}$ represent the position and velocity of the target trajectory in the previous frame, while $\bar T_{v,i}^H$ and $T_{v,i}^H$ represent the velocity of the neighboring trajectory $i$ in the previous and current frames, respectively. $N$ denotes the number of neighboring trajectories. GMCS is capable of aggregating motion priors from neighboring trajectory segments, utilizing the group motion pattern to predict reliable motion states for each target. It performs well in complex scenes.
}

\subsection{Spatio Temporal Memory Prediction}
When the number of neighboring trajectories $N=0$, there are no high-quality trajectories with similar motion states to the low-quality trajectory in the group trajectory. In this case, the group motion compensation strategy fails, and reliable motion state prediction cannot be made using high-quality motion priors. Long Short-Term Memory (LSTM) networks are capable of capturing long-term dependencies in sequences and have advantages in handling sequential data \citep{52yu2019review}. To address the issue of motion prediction for low-quality objects in the absence of neighboring trajectories, we propose a spatio-temporal memory prediction module based on LSTM. By extracting historical trajectory features of the object, the module predicts the object's position at the next time step.
\begin{figure}[!h]
	\centering
	\includegraphics[width=\columnwidth]{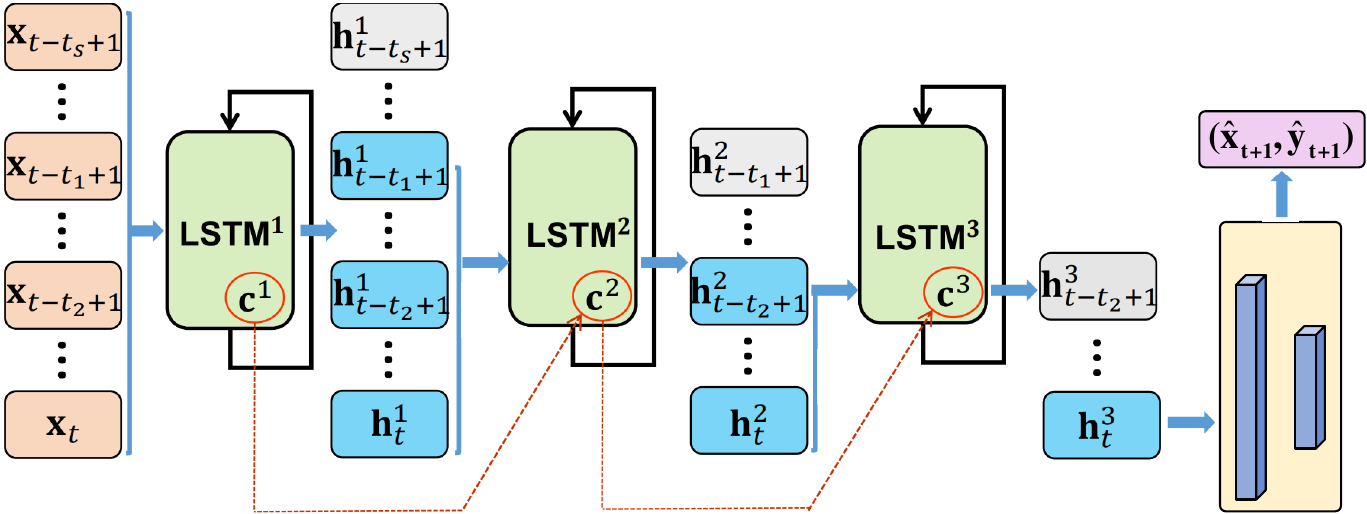}
	\caption{Structure of STMP module.
	}
	\label{fig_4}
\end{figure}

Specifically, STMP consists of a three-stage cascading LSTM network and two fully connected layers, with the time steps of the three LSTMs gradually decreasing to refine the sequence information and focus on the most recent frames, as shown in Fig. \ref{fig_4}. STMP takes the continuous 8-frame position information $({x_{t - n}},{y_{t - n}}),...,({x_{t - 1}},{y_{t - 1}}),({x_t},{y_t})$ as input and outputs the predicted position $({\hat x_{t + 1}},{\hat y_{t + 1}})$ for the next time step. The Mean Squared Error (MSE) loss is chosen as the position prediction loss for network training:
\begin{equation}
	MSE = \frac{1}{n}\sum\nolimits_{i = 1}^n {[{{({x_i} - {{\hat x}_i})}^2} + {{({y_i} - {{\hat y}_i})}^2}]}
\end{equation}

\noindent{where $({x_i},{y_i})$ is the object's center position, and $({\hat x_i},{\hat y_i})$ is the predicted position. We extract all continuous trajectories, crop them into continuous 9-frame sequences, using the first eight frames as training samples and the ninth frame as the label. The optimizer used is Adam, with an initial learning rate of 0.01, and the training is conducted for 100 epochs using a cosine annealing schedule. During the actual inference process, if the object's trajectory does not have 8 continuous frames, STMP is not enabled.}

\section{Experimental Results and Analysis}\label{section_experiment}

In this section, we first provide a description of the dataset, evaluation metrics, and experimental details. Then, detailed and rigorous experiments are conducted on the dataset, comparing our method with baseline methods and state-of-the-art approaches. Furthermore, based on the experimental results, we present a theoretical analysis of the proposed improvements.

\subsection{Datasets}

To better evaluate and validate the effectiveness of ArbiTrack, we choose to conduct a series of experiments using the publicly available UAV tracking dataset UAVDT \citep{53du2018unmanned} and the popular multi-object tracking dataset MOT17 \citep{54milan2016mot16}.

\noindent{\bf{UAVDT:}} UAVDT contains 50 video sequences for multi-object tracking tasks, with 30 sequences for training and 20 sequences for validation. The UAVDT dataset has the following characteristics: 1) High density: wide viewing angles with a large number of targets; 2) Small targets: high shooting altitude with small target sizes; 3) Camera motion: the UAV flies at high speed with camera rotation. The video sequences are stored in JPG format with resolutions of $1080 \times 540$ pixels or $960 \times 540$ pixels and are captured in various complex scenarios, such as plazas, toll stations, highways, and intersections. UAVDT includes three types of natural environments: daytime, nighttime, and foggy conditions, with complex camera viewpoint variations, such as front view, top-down view, and bird's-eye view, which present a significant challenge to multi-object tracking algorithms. UAVDT includes three object categories: cars, trucks, and buses, with a severe class imbalance: cars make up 90\% of all instances, while trucks and buses each account for less than 5\%.

\noindent{\bf{MOT17:}} MOT17 is a classic and challenging dataset for multi-object tracking, focusing on pedestrian multi-object tracking. It contains 14 video sequences, with 7 for training and 7 for testing, at a resolution of $1920 \times 1080$. The main challenges of MOT17 are dense scenes, occlusion, and appearance similarity interference. The image sequences feature dense pedestrians, with severe occlusion and high appearance similarity between objects. We split the training sequences evenly, using the first half for training and the latter half for validation, focusing on the pedestrian category in the multi-object tracking task.

\subsection{UAVDT Correction}

\begin{figure}[!h]
	\centering
	\includegraphics[width=\columnwidth]{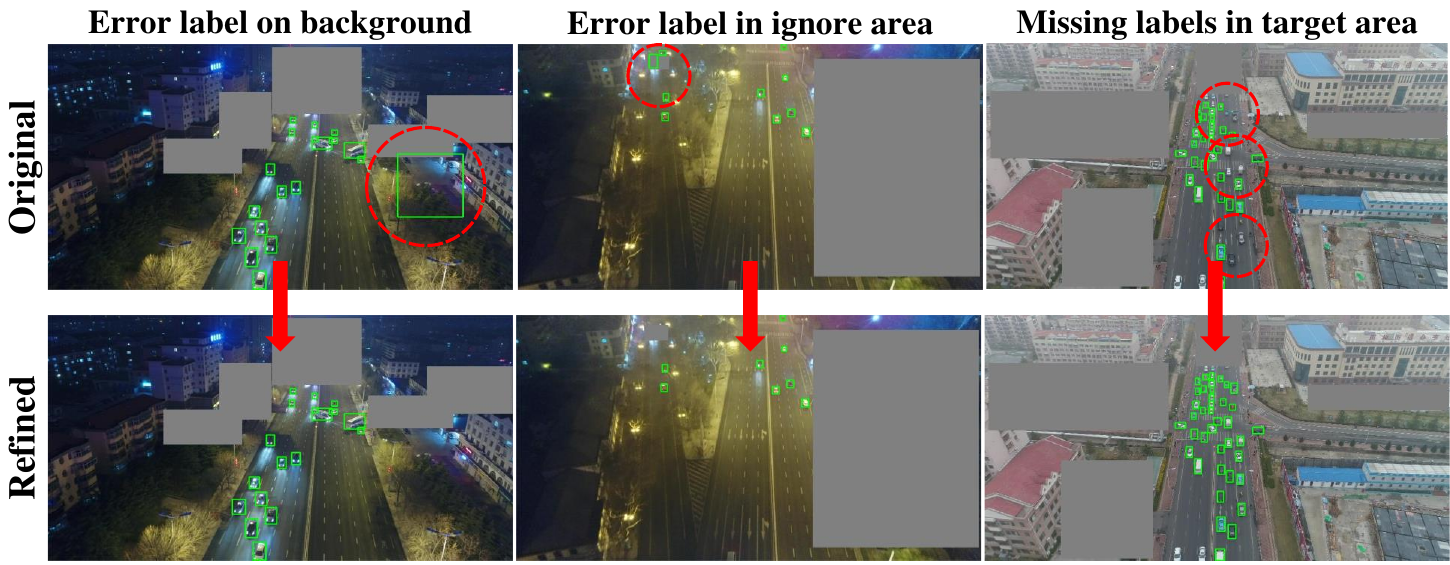}
	\caption{Annotations before and after correction.
	}
	\label{fig_5}
\end{figure}

During the analysis of the dataset, we discovered significant annotation errors in the UAVDT dataset, as shown in Fig. \ref{fig_5}. Notably, some object annotations were completely missing, while some objects were located in ignored areas but their corresponding annotations were not removed. Additionally, there were some instances where large-sized labels were incorrectly placed on backgrounds without any objects. To mitigate the negative impact of these erroneous annotations on the experiments, we performed a comprehensive cleaning of the annotation files.

\subsection{Evaluation Metrics}
We evaluate the tracking performance using the CLEAR metric system \citep{55bernardin2008evaluating}, HOTA \citep{56luiten2021hota}, MT, and ML. The CLEAR metric system includes MOTA, MOTP, IDF1, IDSW, etc. MOTA comprehensively measures the overall accuracy of the tracker in terms of detection errors and ID preservation:
\begin{equation}
	{\rm{MOTA}} = 1 - \frac{{({\rm{FN}} + {\rm{FP}} + {\rm{IDSW}})}}{{{\rm{GT}}}}
\end{equation}

\noindent{where $GT$ is the number of ground-truth targets. MOTP measures the localization accuracy between the tracking bounding boxes and the ground-truth boxes:}
\begin{equation}
	MOTP = \frac{{{\rm{IoU}}({\rm{Track}},{\rm{GT}})}}{{{\rm{Matches}}}}
\end{equation}

\noindent{where $Matches$ represents the successfully matched trajectories. IDF1 balances the precision and recall of ID preservation:}
\begin{equation}
	IDF1 = \frac{{2 \times IDTP}}{{2 \times IDTP + IDFP + IDFN}}
\end{equation}

HOTA provides a balanced evaluation of the comprehensive efficiency of detection and association. MT and ML assess the stability and continuity of the tracking algorithm. Specifically, MT refers to the number of trajectories where the number of successfully tracked frames accounts for more than 80\% of the total length of the trajectory, while ML refers to the number of trajectories where the number of successfully tracked frames accounts for less than 20\% of the total length of the trajectory.

\subsection{Implementation Details}

We implemented the proposed tracking algorithm framework based on PyTorch, and all experiments were conducted on a server platform equipped with four NVIDIA 4090 GPUs using the same configuration. For UAVDT, we trained the detector for 300 epochs with a batch size of 32. The optimizer used was Stochastic Gradient Descent (SGD), with the learning rate decaying from 0.01 to 0.0001 using a cosine annealing schedule. For MOT17, to ensure a fair comparison, we directly used the pre-trained YOLOX \citep{57ge2021yolox} detector to generate the target detection results. Unless otherwise specified, all experiments maintained the above configuration.

\subsection{Comparison with Other State-of-the-Art Methods}

To comprehensively evaluate the overall performance of the proposed SocialTrack algorithm in multi-object tracking tasks, we conducted a systematic comparison with current mainstream multi-object tracking methods on the challenging UAVDT test set. The comparison included traditional trackers (such as SORT, DeepSORT), end-to-end methods (such as JDE, FairMOT), and recently proposed advanced algorithms (such as OC-SORT, ByteTrack, BoT-SORT). The comparison results are shown in Table \ref{table_1}. From the results, it can be seen that SocialTrack significantly outperforms existing methods on several key evaluation metrics, achieving the current best performance.
\begin{table}[!h]
	\renewcommand{\arraystretch}{1.2}
	\centering
	\caption{Comparison of SocialTrack with other methods on the UAVDT dataset. The last row shows our proposed SocialTrack, which obtained SOTA performance in almost all categories.}
	\label{table_1}  
	\setlength{\tabcolsep}{2pt}  
	\begin{tabular*}{\textwidth}{lcccccccc} 
		\toprule[1pt]
		\textbf{Method} & \textbf{MOTA$\uparrow$} & \textbf{MOTP$\uparrow$} & \textbf{IDF1$\uparrow$} & \textbf{MT$\uparrow$} & \textbf{ML$\downarrow$} & \textbf{FP$\downarrow$} & \textbf{FN$\downarrow$} & \textbf{IDSW$\downarrow$} \\
		\midrule[0.75pt]
		SMOT \citep{58dicle2013way} & 33.9 & 72.2 & 45 & 524 & 367 & 57112 & 166528 & 1752 \\
		IOUT \citep{59bochinski2017high} & 36.6 & 72.1 & 32.7 & 534 & 357 & 42245 & 163881 & 993 \\
		CMOT \citep{60bae2014robust} & 36.9 & 74.7 & 57.5 & 664 & 351 & 69109 & 144760 & 1111 \\
		JDE \citep{34tsai2023swin} & 39.5 & 73.5 & 55.3 & 624 & 442 & - & - & 3124 \\
		FairMOT \citep{35zhang2021fairmot} & 44.9 & 72.7 & 66.5 & 665 & 326 & - & - & 2279 \\
		MDP \citep{61xiang2015learning} & 43 & 73.5 & 61.5 & 647 & 324 & 46151 & 147735 & 541 \\
		SuperMOT \citep{2ren2025supermot} & 50.6 & 75.7 & 70.6 & 674 & 227 & 60587 & 107538 & 407 \\
		MOTDT \citep{46chen2018real} & 36.4 & - & 56.4 & 464 & 402 & 41505 & 196014 & 590 \\
		SORT \citep{23bewley2016simple} & 39 & 74.3 & 43.7 & 484 & 400 & 33037 & 172628 & 2350 \\
		DeepSORT \citep{25wojke2017simple} & 40.7 & 73.2 & 52.2 & 595 & 358 & 44868 & 155290 & 2061 \\
		OC-SORT \citep{22cao2023observation} & 52.1 & - & 69.8 & 535 & 192 & 37908 & 141340 & 286 \\
		BoT-SORT \citep{21aharon2022bot} & 52.5 & - & 69.7 & 646 & 201 & 56067 & 121304 & 441 \\
		ByteTrack \citep{26zhang2022bytetrack} & 53 & - & 70.3 & 668 & 193 & 56021 & 119674 & \textbf{215} \\
		SFTrack \citep{62song2024sftrack} & 55.3 & - & 71.3 & 682 & 184 & 49946 & 117224 & 284 \\
		\midrule[0.5pt]
		SocialTrack (ours) & \textbf{64.6} & \textbf{76.3} & \textbf{76.1} & \textbf{702} & \textbf{170} & \textbf{31133} & \textbf{89146} & 276 \\
		\bottomrule[1pt]
	\end{tabular*}
\end{table}

\begin{figure*}[!h]
	\centering
	\includegraphics[width=\textwidth]{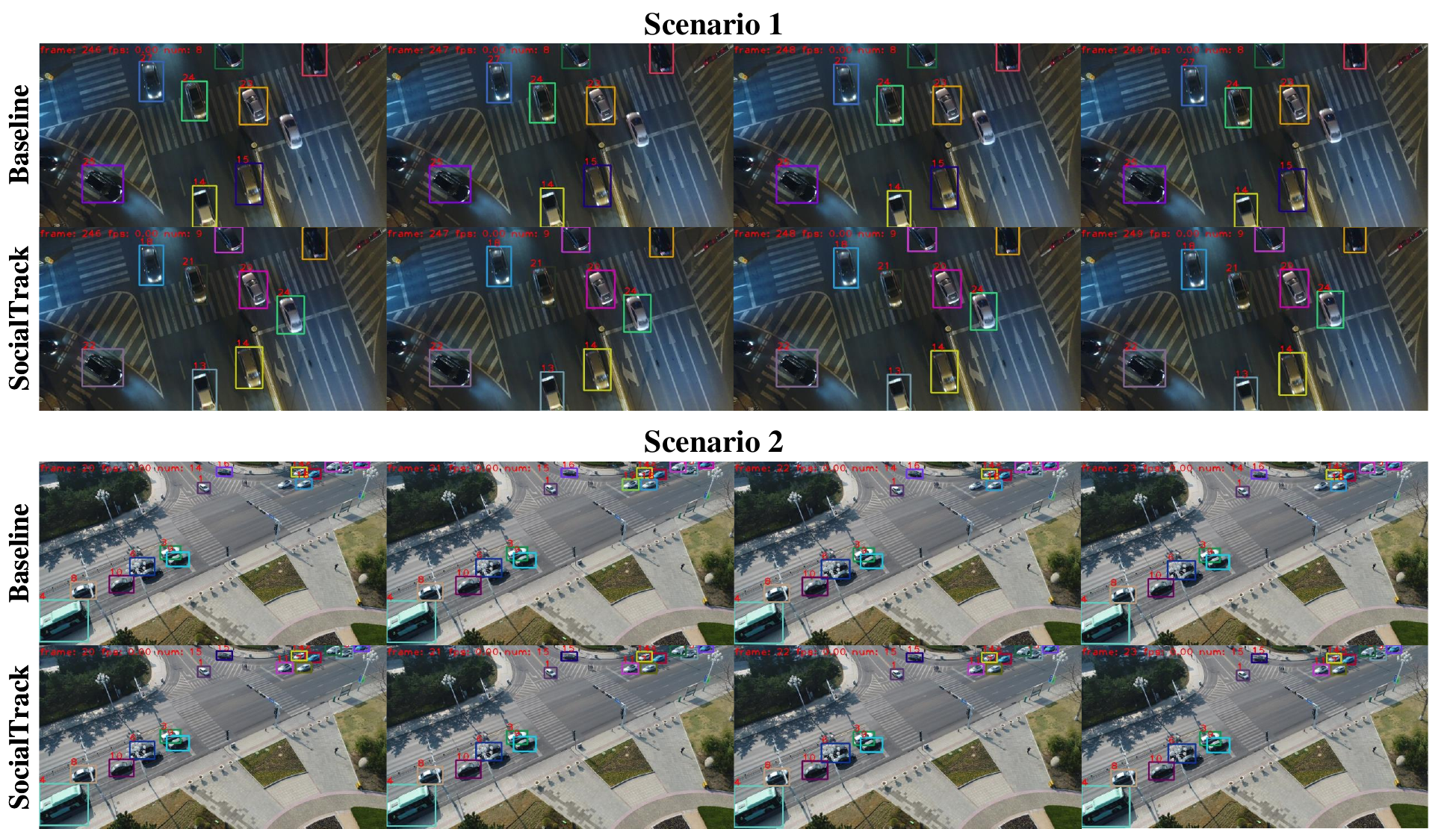}
	\caption{Visual tracking results of the baseline method and the proposed SocialTrack in different scenarios in the UAVDT dataset.
	}
	\label{fig_6}
\end{figure*}

Specifically, on the MOTA metric, SocialTrack achieved 64.6\%, which is an improvement of 9.3 and 11.6 percentage points compared to the current best methods, SFTrack (55.3\%) and ByteTrack (53.0\%), respectively. This indicates that the multi-module fusion strategy we proposed significantly enhances the overall performance of target detection, data association, and state estimation. On the MOTP metric, SocialTrack ranked first with a result of 76.3\%, demonstrating strong performance in target localization accuracy. Additionally, on the crucial identity consistency metric, IDF1, SocialTrack achieved 76.1\%, significantly outperforming ByteTrack (70.3\%), SuperMOT (70.6\%), and SFTrack (71.3\%), showing that the proposed method can maintain stable tracking of target identities even in complex situations involving occlusion and multi-target crossing. At the same time, on the MT and ML metrics, SocialTrack achieved 702 and 170, respectively, both the best among all methods, indicating that the system can track most targets stably and significantly reduce long-term target loss. In terms of FP and FN, SocialTrack achieved 31,133 and 89,146, respectively, significantly outperforming other two-stage trackers, confirming its effective suppression of low-quality detection results, further ensuring overall tracking accuracy and robustness. At the same time, the IDSW metric was reduced to 276, which is slightly lower than the optimal ByteTrack (215), but further optimized compared to OC-SORT (286) and SFTrack (284), fully demonstrating the effectiveness of mechanisms like spatio-temporal memory prediction and group motion compensation in reducing identity switching. Fig. \ref{fig_6} presents the tracking results between the baseline method and the proposed SocialTrack across various scenarios in the UAVDT dataset.

\subsection{Ablation Experiments}

We conducted a series of ablation experiments on the UAVDT and MOT17 validation sets to verify the effectiveness and reliability of the proposed method.

\noindent{\bf{UAVDT:}} In this study, ByteTrack \citep{26zhang2022bytetrack} was used as the baseline, and the effectiveness of each proposed module was validated through systematic ablation experiments. Table \ref{table_2} shows the improvements in overall tracking performance due to four key modifications. First, to enhance the detector's ability to perceive small targets, we designed a specialized detector for efficient small-target feature perception. The experimental results indicate that this module improved MOTA from 56.1 to 62.5, MOTP from 75.2 to 76.0, and IDF1 from 68.6 to 74.1, with significant progress in all three key metrics. This demonstrates that the improvement in detection quality plays a crucial role in the overall tracking performance. Additionally, IDSW decreased by 128, indicating that more accurate target detection also helps maintain identity consistency in the subsequent association process. Building on the detection results, we introduced the VACKF, which combines the advantages of the velocity-adaptive model and cubature Kalman filter for nonlinear state modeling and prediction. This module increased MOTA to 63.2, MOTP slightly to 76.1, and IDF1 to 74.7, while further decreasing IDSW to 304. This result demonstrates that the adaptive filtering mechanism significantly reduces trajectory estimation errors and identity switching frequency in scenes with frequent occlusions or complex motion, thereby enhancing the robustness and continuity of the tracking system. Next, we introduced the GMCS, which leverages the motion prior information from neighboring trajectories in the group to provide robust relative state references for low-quality trajectories, further improving the accuracy of their state estimation. This module increased MOTA to 64.2, IDF1 to 75.4, and reduced IDSW to 286, confirming the value of group behavior information in weak observation scenarios. Finally, when the GMCS fails, the STMP uses historical trajectory information to predict the future state of low-quality trajectories, effectively enhancing the temporal robustness of the model. This module further increased MOTA to 64.6, IDF1 to 76.1, and reduced IDSW to 276, indicating its ability to maintain target identity continuity in complex scenes.
\begin{table}[!h]
	\renewcommand{\arraystretch}{1.5}
	\centering
	\caption{Complete results of the ablation study on the UAVDT dataset.}
	\label{table_2}
		\setlength{\tabcolsep}{4pt}  
		\begin{tabular*}{\textwidth}{lcccccc}
			\toprule[1pt]
			\textbf{Method} & \textbf{MOTA$\uparrow$} & \textbf{MOTP$\uparrow$} & \textbf{IDF1$\uparrow$} & \textbf{IDSW$\downarrow$} & \textbf{MT$\uparrow$} & \textbf{ML$\downarrow$} \\
			\midrule[0.75pt]
			Baseline & 56.1 & 75.2 & 68.6 & 447 & 442 & 196 \\
			+SOFEPNet & 62.5 & 76.0 & 74.1 & 319 & 645 & 174 \\
			+SOFEPNet+VACKF & 63.2 & 76.1 & 74.7 & 304 & 663 & 170 \\
			+SOFEPNet+VACKF+GMCS & 64.2 & 76.2 & 75.4 & 286 & 689 & \textbf{169} \\
			+SOFEPNet+VACKF+GMCS+STMP & \textbf{64.6} & \textbf{76.3} & \textbf{76.1} & \textbf{276} & \textbf{702} & 170 \\
			\bottomrule[1pt]
		\end{tabular*}
\end{table}

To address the limitations of classical Kalman filter in predicting nonlinear motion, we introduced several enhancements to the filtering process: a velocity-adaptive model was employed to capture the nonlinear dynamics of moving objects and was integrated into the CKF. As shown in Table \ref{table_7}, the inclusion of the velocity-adaptive model led to an increase in MOTA to 62.7 and IDF1 to 74.4, indicating the model's ability to adapt to non-uniform motion and effectively mitigate estimation biases caused by conventional linear prediction assumptions. Building on this, the integration of the CKF resulted in the complete VACKF model, which further improved MOTA to 63.2 and IDF1 to 74.7, representing the highest scores in this group of experiments. These results demonstrate that CKF's capability to model nonlinear state transitions complements the adaptive motion modeling, enhancing the filter's ability to cope with complex and dynamic scenarios. Overall, the experimental results confirm that the design of VACKF significantly improves state estimation accuracy, reduces trajectory jitter and identity switches, and enhances the overall stability and robustness of the tracking system under challenging conditions.
\begin{table}[h]
	\renewcommand{\arraystretch}{1.5} 
	\caption{Performance test of VACKF module.}
	\label{table_7}
	\centering
		\setlength{\tabcolsep}{8pt}  
		\begin{tabular}{c | c c | c c}
			\toprule[1pt]
			& \textbf{Speed Adaptive Model} & \textbf{CKF} & \textbf{MOTA $\uparrow$} & \textbf{IDF1 $\uparrow$}\\
			\midrule[0.75pt]
			\multirow{3}{*}{\textbf{VACKF}} & & & 62.5 & 74.1 \\
			& \checkmark & & 62.7 & 74.4 \\
			& \checkmark & \checkmark & \textbf{63.2} & \textbf{74.7} \\
			\bottomrule[1pt]
		\end{tabular}
\end{table}

\noindent{\bf{MOT17:}} To further validate the generalization and robustness of the proposed modules, we integrated SocialTrack (excluding the specialized detector) into two mainstream multi-object tracking frameworks--ByteTrack \citep{26zhang2022bytetrack} and BoT-SORT \citep{21aharon2022bot}--and conducted systematic ablation experiments on the MOT17 dataset while keeping the detector consistent (YOLOX-X). The experimental results are shown in Table \ref{table_3} and Table \ref{table_4}.
\begin{table}[h]
	\renewcommand{\arraystretch}{1.5}
	\centering
	\caption{Performance improvement with different components added to ByteTrack.}
	\label{table_3}
		\setlength{\tabcolsep}{5pt}  
		\begin{tabular*}{\textwidth}{lcccccc}
			\toprule[1pt]
			\textbf{Method} & \textbf{HOTA$\uparrow$} & \textbf{MOTA$\uparrow$} & \textbf{IDF1$\uparrow$} & \textbf{IDR$\uparrow$} & \textbf{IDP$\uparrow$} & \textbf{IDSW$\downarrow$} \\
			\midrule[0.75pt]
			ByteTrack & 68.0 & 76.6 & 79.3 & 75.1 & 84.1 & 159 \\
			ByteTrack+VACKF & 68.1 & \textbf{76.7} & 79.7 & 75.4 & 84.4 & 160 \\
			ByteTrack+VACKF+GMCS & \textbf{68.4} & \textbf{76.7} & 80.2 & 75.9 & 85.0 & 156 \\
			ByteTrack+VACKF+GMCS+STMP & 68.3 & \textbf{76.7} & \textbf{80.4} & \textbf{76.1} & \textbf{85.3} & \textbf{139} \\
			\bottomrule[1pt]
		\end{tabular*}
\end{table}
\begin{figure*}[!h]
	\centering
	\includegraphics[width=\textwidth]{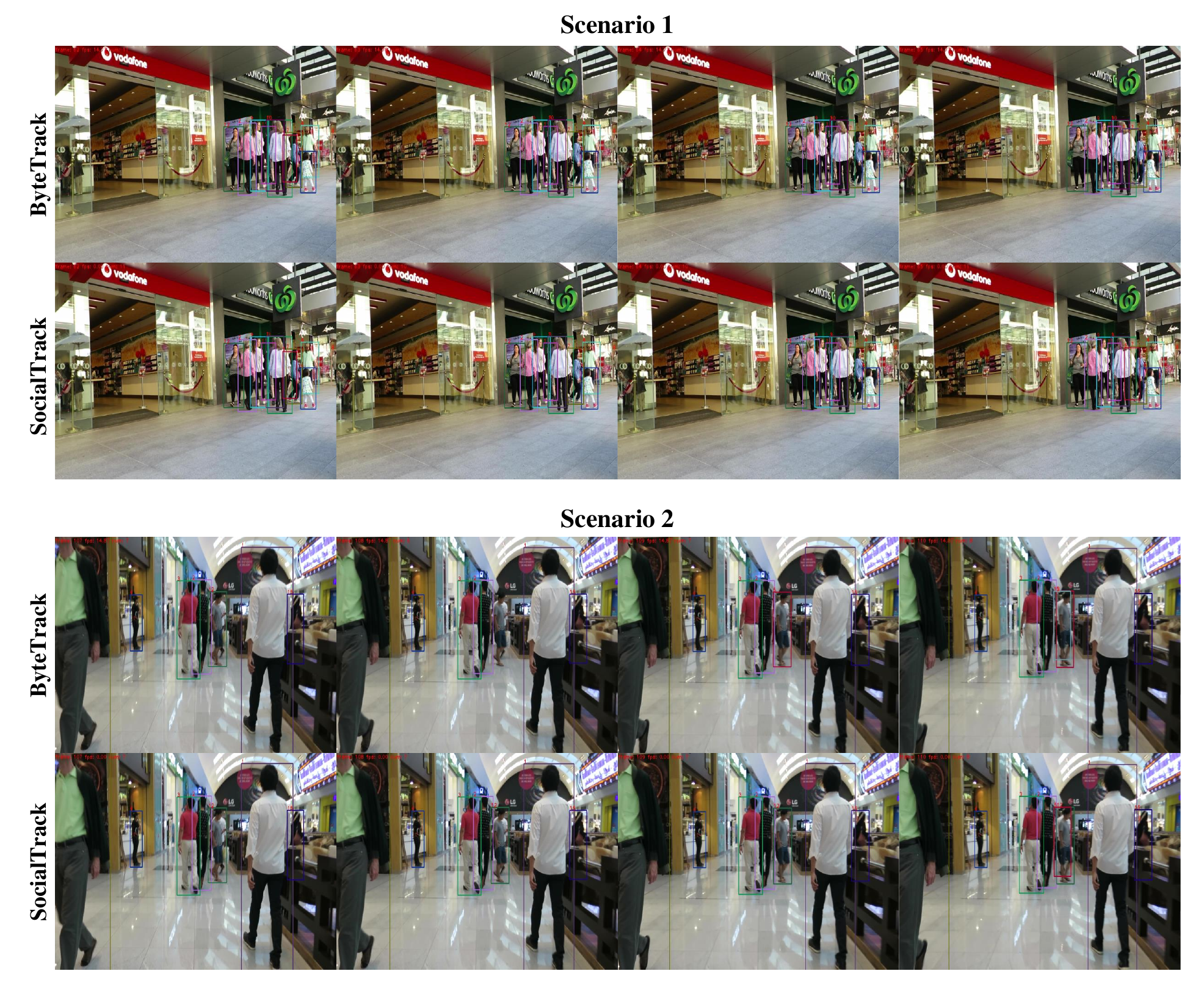}
	\caption{Visual tracking results of the ByteTrack and the proposed SocialTrack in different scenarios in the MOT17 dataset.
	}
	\label{fig_7}
\end{figure*}

In the ByteTrack framework, after adding VACKF, HOTA improved from 68.0 to 68.1, MOTA increased from 76.6 to 76.7, IDF1 rose from 79.3 to 79.7, and both IDR and IDP showed slight increases. This indicates that even under conditions of limited detection accuracy, VACKF can enhance identity preservation and overall stability in the tracking process through its nonlinear modeling capabilities and adaptive state estimation mechanism. After introducing the GMCS, the overall performance improved further, with IDF1 reaching 80.2, HOTA rising to 68.4, and IDSW decreasing from 159 to 156. By effectively modeling the motion consistency among individuals in the local neighborhood, this module demonstrated stronger identity continuity in scenarios involving occlusion and crossing motion. With the addition of the STMP, IDF1 further increased to 80.4, and IDSW significantly dropped to 139, demonstrating that this module can accurately predict states in situations with low observation quality or temporary target loss, thereby effectively mitigating identity switching and enhancing the system's temporal robustness. Fig. \ref{fig_7} presents the tracking results between the ByteTrack and the proposed SocialTrack across various scenarios in the MOT17 dataset.
\begin{table}[!h]
	\renewcommand{\arraystretch}{1.5}
	\centering
	\caption{Performance improvement with different components added to BoT-SORT.}
	\label{table_4}
		\setlength{\tabcolsep}{4pt}  
		\begin{tabular*}{\textwidth}{lcccccc}
			\toprule[1pt]
			\textbf{Method} & \textbf{HOTA$\uparrow$} & \textbf{MOTA$\uparrow$} & \textbf{IDF1$\uparrow$} & \textbf{IDR$\uparrow$} & \textbf{IDP$\uparrow$} & \textbf{IDSW$\downarrow$} \\
			\midrule[0.75pt]
			BoT-SORT & 69.1 & 78.3 & 81.4 & 76.9 & 86.5 & 140 \\
			BoT-SORT+VACKF & 69.1 & \textbf{78.4} & 81.4 & 76.9 & 86.5 & 140 \\
			BoT-SORT+VACKF+GMCS & 69.3 & \textbf{78.4} & 81.7 & 76.9 & 87.0 & 123 \\
			BoT-SORT+VACKF+GMCS+STMP & \textbf{69.4} & \textbf{78.4} & \textbf{82.2} & \textbf{77.4} & \textbf{87.6} & \textbf{122} \\
			\bottomrule[1pt]
		\end{tabular*}
\end{table}
\begin{figure*}[!h]
	\centering
	\includegraphics[width=\textwidth]{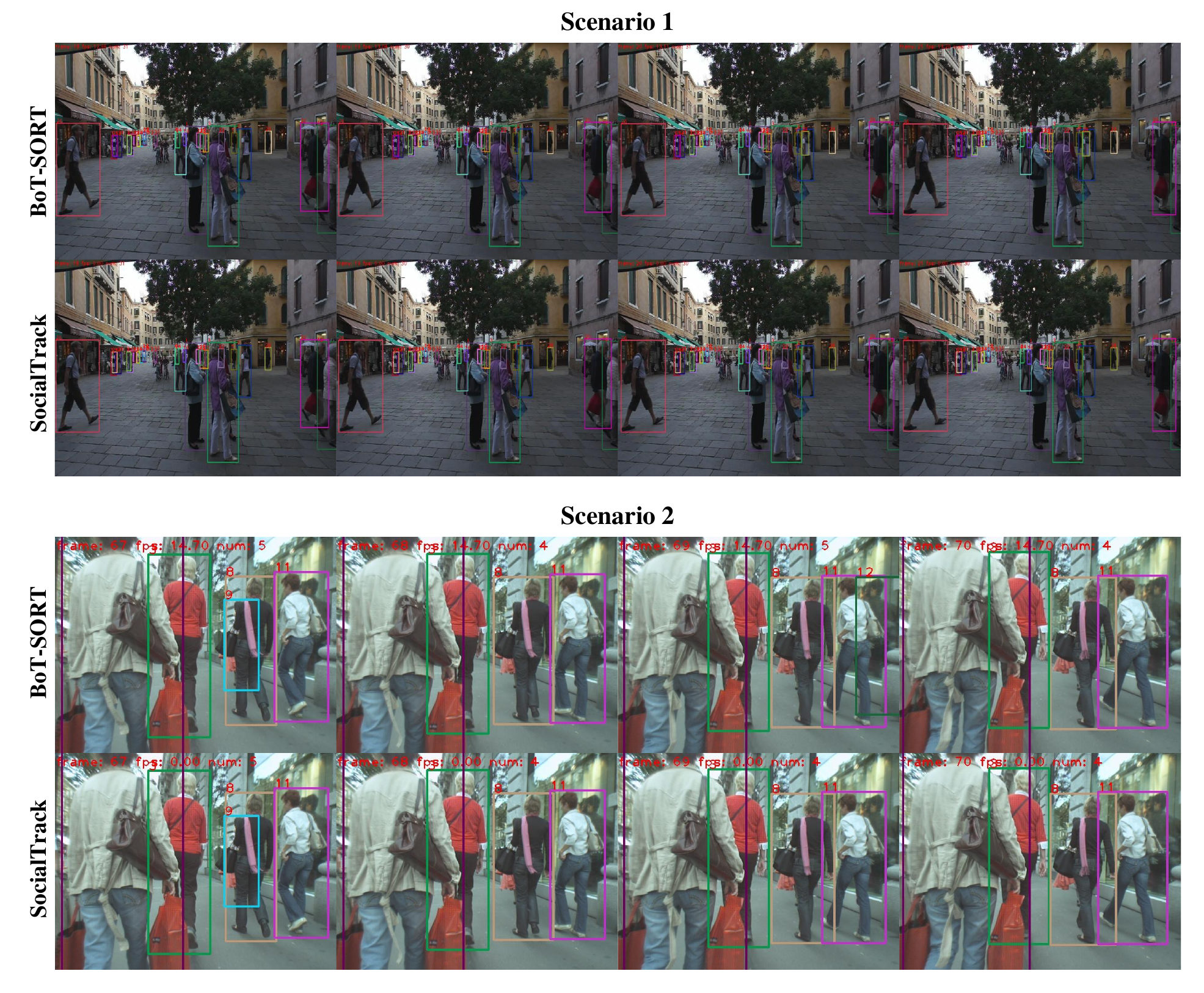}
	\caption{Visual tracking results of the BoT-SORT and the proposed SocialTrack in different scenarios in the MOT17 dataset.
	}
	\label{fig_8}
\end{figure*}

In the BoT-SORT framework, a similar trend was observed, further confirming the good generalizability of the proposed method. Compared to the original BoT-SORT, VACKF did not lead to performance improvements, which can be attributed to the camera motion compensation already addressing the position prediction shortcomings. After introducing the GMCS, IDF1 improved from 81.4 to 81.7, IDP increased to 87.0, and IDSW significantly decreased to 123, indicating that this strategy can effectively integrate motion prior information into BoT-SORT's feature modeling mechanism, improving association accuracy in weak observation scenarios. Finally, with the addition of the STMP, the IDF1 of BoT-SORT increased to 82.2, HOTA rose to 69.4, and IDSW further decreased to 122, showing a consistent performance improvement trend similar to ByteTrack. This module demonstrated robust and transferable performance enhancement across different frameworks, further proving the proposed method's good cross-framework adaptability and potential for broad application. Fig. \ref{fig_8} presents the tracking results between the BoT-SORT and the proposed SocialTrack across various scenarios in the MOT17 dataset.

\section{Conclusion}\label{conclusion}

This study proposes a novel multi-object tracking algorithm, SocialTrack, designed to address the challenges of tracking small targets in complex environments. Through a series of ablation and comparison experiments, we have demonstrated the effectiveness of the proposed improvements in enhancing tracking accuracy, robustness, and stability. Compared to current mainstream methods, SocialTrack leads in several metrics, particularly in core performance indicators such as MOTA (64.6\%) and IDF1 (76.1\%), achieving improvements of 9.3\% and 4.8\%, respectively, over the existing best methods, setting a new state-of-the-art. Additionally, SocialTrack shows superior performance in handling complex scenarios, particularly under frequent occlusions and dense targets, effectively reducing identity switching and target loss, demonstrating stronger robustness.

In conclusion, SocialTrack significantly improves the overall performance of multi-object tracking in complex environments by systematically integrating multiple innovative modules. Furthermore, all innovative modules possess strong plug-and-play capabilities and compatibility, allowing them to be incorporated into existing tracking algorithms to further enhance performance, demonstrating the great potential of this method for practical applications. Therefore, SocialTrack provides an important reference and insight for future multi-object tracking research and engineering applications.


\section*{Declarations}
\bmhead{Acknowledgements}
This research received funding from the National Natural Science Foundation of China under Grant 62106193 and from the Fundamental Research Funds for the Central Universities under Grant G2024KY05105.

\bmhead{Author contributions}
Wenguang Tao: Data Curation, Conceptualization, Methodology, Software, Validation, Writing-Original Draft. Xiaotian Wang: Conceptualization, Software, Validation. Tian Yan \& Guodong Li: Software, Validation, Visualization. Jie Yan: Supervision, Validation, Writing-Review \& Editing. Kun Bai: Formal analysis, Supervision, Validation, Writing-Review \& Editing. All authors reviewed the manuscript.

\bmhead{Data availability}
No datasets were generated or analysed during the current study.

\bmhead{Conflict of interest}
The authors declare no competing interests.


\bibliography{sn-bibliography}

\end{document}